\documentclass{amsart}
\usepackage{amsaddr}
\usepackage{graphicx}
\usepackage{hyperref}

\begin{document}

\title{Evolving Evolutionary Algorithms 
\\using Multi Expression Programming}

\author{Mihai Oltean
\and Crina Gro\c san}

\address{Department of Computer Science,\\
Faculty of Mathematics and Computer Science,\\
Babes-Bolyai University, Kogalniceanu 1,\\
Cluj-Napoca, 3400, Romania.\\
\url{https://mihaioltean.github.io}
}
\email{mihai.oltean@gmail.com}

\maketitle

\maketitle

\begin{abstract}
Finding the optimal parameter setting (i.e. the optimal population size, the optimal mutation probability, the optimal evolutionary model etc) for an Evolutionary Algorithm (EA) is a difficult task. Instead of evolving only the parameters of the algorithm we will  evolve an entire EA capable of solving a particular problem. For this purpose the Multi Expression Programming (MEP) technique is used. Each MEP chromosome will encode multiple EAs. An nongenerational EA for function optimization is evolved in this paper. Numerical experiments show the effectiveness of this approach.
\end{abstract}

\section{Introduction}

Evolutionary Algorithms (EAs) \cite{Goldberg, Holland} are nonconventional tools for solving difficult real-world problems. They were developed under the pressure generated by the inability of classical (mathematical) methods to solve some real-world problems. Many of these unsolved problems are (or could be turned into) optimization problems. Solving an optimization problem means finding of solutions that maximize or minimize a criteria function \cite{Goldberg}.

Many EAs were proposed for dealing with optimization problems. Many solution representations and search operators were proposed and tested within a wide range of evolutionary models. There are several natural questions that are to be answered in all of these evolutionary models: which is the optimal population size?, which is the optimal individual representation?, which are the optimal probabilities for applying specific genetic operators?, which is the optimal number of generations before halting the evolution? etc.

A breakthrough arose in 1995 when Wolpert and McReady unveiled their work on the No Free Lunch (NFL) theorems \cite{Wolpert}. The NFL theorems state that all of the black-box algorithms perform equally well over the entire set of optimization problems. (A black-box algorithm does not take into account any information about the problem or the particular instance being solved.) The magnitudes of the NFL results stroke all of the efforts for developing a universal black-box optimization algorithm able to solve best all the optimization problems.

In their attempt to solve problems, men delegated computers to develop algorithms able to perform certain tasks. The most prominent effort in this direction is Genetic Programming (GP) \cite{Koza}, an evolutionary technique used for breeding a population of computer programs. Instead of evolving solutions for a particular problem instance, GP is mainly intended for discovering computer programs able to solve particular classes of problems. (This statement is only partially true, since the discovery of computer programs may be also viewed as a technique for solving a particular problem instance. The following could be an example of a problem: 'Find a computer program that calculates the sum of the elements of an array of integers.')

There are many such approaches so far in the GP literature \cite{Brameier}. The evolving of deterministic computer programs able to solve specific problems requires a lot of effort.

Instead of evolving deterministic computer programs we will try to evolve a full-featured evolutionary algorithm (i.e. the output of our main program will be an EA able to perform a given task). Thus we will work with EAs at two levels: the first (macro) level consists of a steady-state EA \cite{Syswerda} which uses a fixed population size, a fixed mutation probability, a fixed crossover probability etc. The second (micro) level consists of the solution encoded in a chromosome from the GA on the first level. 

Having this aim in view we use an evolutionary model similar to Multi Expression Programming (MEP) \cite{Oltean} \footnote{MEP source code is available at \url{https://mepx.org} and \url{https://mepx.github.io}} which is very suitable for evolving computer programs that may be easily translated into an imperative language (like C or Pascal). The evolved EA is a nongenerational one (i.e. there is no cycle during evolution).

The paper is organized as follows: the MEP technique is described in section \ref{mep_s}. The model used for evolving EAs is presented in section \ref{model_s}. The way in which the fitness of an MEP individual is computed is described in section \ref{fitness_s}. Several numerical experiments are performed in section \ref{experiments_s}.

\section {MEP technique}
\label{mep_s}

The Multi Expression Programming technique is briefly described in this section. 

\subsection{The MEP Algorithm}
\label{sect:The MEP Algorithm_online}

In this paper we use steady-state \cite{Syswerda} as the underlying mechanism for MEP.
The steady-state MEP algorithm starts with a randomly chosen population of individuals. The following steps are repeated until a termination condition is reached: Two parents are selected (out of 4 individuals) by using a binary tournament procedure \cite{Koza} and they are recombined with a fixed crossover probability. By the recombination of two parents two, offspring are obtained. The offspring are mutated and the best of them replaces the worst individual in the current population (if the offspring is better than the worst individual in the current population).

\subsection{The MEP representation}
\label{sect:The MEP representation_online}

The MEP genes are (represented by) substrings of variable length. The number of genes in a chromosome is constant and it represents the chromosome length. Each gene encodes a terminal (an element in the terminal set $T$) or a function symbol (an element in the function set $F$). A gene encoding a function includes pointers towards the function arguments. Function parameters always have indices of lower values than the position of that function itself in the chromosome. 

According to the proposed representation scheme, the first symbol in a chromosome must be a terminal symbol. In this way only syntactically correct programs are obtained.
\\
\\
{\bf Example}

We use a representation where the numbers on the left positions stand for gene labels (or memory addresses). Labels do not belong to the chromosome, they are provided only for explanatory purposes.
An example of a chromosome is given below (assuming that $T$ = \{$a$, $b$, $c$, $d$\} and $F$ = \{+, -, *, /\}):
\\
1: $a$\\
2: $b$\\
3: + 1, 2\\
4: $c$\\
5: $d$\\
6: + 4, 5\\
7: * 2, 6\\
\subsection{The MEP phenotypic transcription}

This section is devoted to describing the way in which the MEP individuals are translated into computer programs.

The MEP chromosomes are read in a top-down fashion starting with the first position. A terminal symbol specifies a simple expression. A function symbol specifies a complex expression (formed by linking the operands specified by the argument positions with the current function symbol). 

For instance, genes 1, 2, 4 and 5 in the previous example encode simple expressions formed by a single terminal symbol. These expressions are: $E_1 = a$; $E_2 = b$; $E_4 = c$; $E_5 = d$.

Gene 3 indicates the operation + on the operands located at positions 1 and 2 of the chromosome. Therefore gene 3 encodes the expression: $E_3 = a + b$.

Gene 6 indicates the operation + on the operands located at positions 4 and 5. Therefore gene 6 encodes the expression: $E_6 = c + d$.

Gene 7 indicates the operation * on the operands located at positions 2 and 6. Therefore gene 7 encodes the expression: $E_3 = b * (c + d)$.

We have to choose one of these expressions ($E_1,\dots,E_7$) to represent the chromosome. There is neither theoretical nor practical evidence that one of them is better than the others. Thus, we choose to encode multiple solutions in a single chromosome. Each MEP chromosome encodes a number of expressions equal to the chromosome length (the number of genes). The expression associated to each chromosome position is obtained by reading the chromosome bottom-up from the current position, by following the links provided by the functions pointers. The fitness of each expression encoded in a MEP chromosome is computed in a conventional manner (the fitness depends on the problem being solved). The best expression encoded in a MEP chromosome is chosen to represent the chromosome (the fitness of a MEP individual equals the fitness of the best expression encoded in that chromosome).

\section{The Evolutionary Model}
\label{model_s}

In order to use MEP for evolving EAs we have to define a set of terminal symbols and a set of function symbols. When we define these sets we have to keep in mind that the value stored by a terminal symbol is independent of other symbols in the chromosome and a function symbol changes the solution stored in another gene. 

An EA usually has 4 types of genetic operators:
\begin{itemize}
\item {\it Initialize} - randomly initializes a solution,
\item {\it Select} - selects the best solution among several already existing solutions
\item {\it Crossover} - recombines two already existing solutions,
\item {\it Mutate} - varies an already existing solution.
\end{itemize}

These operators will act as symbols that may appear into an MEP chromosome. The only operator that generates a solution independent of the already existing solutions is the {\it Initialize} operator. This operator will constitute the terminal set. The other operators will be considered function symbols. Thus, we have $T$ = \{{\it Initialize}\}, $F$ = \{{\it Select}, {\it Crossover}, {\it Mutate}\}.

A MEP chromosome $C$, storing an evolutionary algorithm is:
\\
\\
{\tt
1: {\it Initialize}\hspace{1.88cm}\{Randomly generates a solution.\}\\
2: {\it Initialize}\hspace{1.88cm}\{Randomly generates another solution.\}\\
3: {\it Mutate} 1\hspace{1,70cm}\{Mutates the solution stored on position 1\}\\
4: {\it Select} 1, 3\hspace{1,41cm}\{Selects the best solution from those\}\\
\hspace*{3,82cm}\{stored on positions 1 and 3\}\\
5: {\it Crossover} 2, 4\hspace{0,75cm}\{Recombines the solutions on positions 2 and 4\}\\ 
6: {\it Mutate} 4\hspace{1,79cm}\{Mutates the solution stored on position 4\}\\
7: {\it Mutate} 5\hspace{1,79cm}\{Mutates the solution stored on position 5\}\\
8: {\it Crossover} 2, 6\hspace{0,79cm}\{Recombines the solutions on positions 2 and 6\}\\
}

This MEP chromosome encodes multiple evolutionary algorithms. Each EA is obtained by reading the chromosome bottom up, starting with the current gene and following the links provided by the function pointers. Thus we deal with EAs at two different levels: a micro level representing the evolutionary algorithm encoded in a MEP chromosome and a macro level GA, which evolves MEP individuals. The number of genetic operators (initializations, crossovers, mutations, selections) is not fixed and it may vary between 1 and the MEP chromosome length. These values are automatically discovered by the evolution. The macro level GA execution is bound by the known rules for GAs (see \cite{Goldberg}).

For instance, the chromosome defined above encodes 8 EAs. They are given in Table 1.
\setlength{\tabcolsep}{4pt}
\begin{table}
\begin{center}
\caption{Evolutionary Algorithms encoded in the MEP chromosome $C$}
\label{table:headings}
\begin{tabular}{|l|l|l|l|}
\hline
$EA_1$ & $EA_2$ & $EA_3$ &	$EA_4$\\
\hline
$i_1$={\it Initialize} & $i_1$={\it Initialize} & $i_1$={\it Initialize} & $i_1$ = {\it Initialize}\\
 & & $i_2$={\it Mutate}($i_1$) & $i_2$={\it Mutate}($i_1$)\\
 & & & $i_3$={\it Select}($i_1$,$i_2$)\\
\hline
$EA_5$ & $EA_6$ & $EA_7$ & $EA_8$\\
\hline
$i_1$={\it Initialize} & $i_1$={\it Initialize} & $i_1$={\it Initialize} & $i_1$={\it Initialize}\\
$i_2$={\it Initialize} & $i_2$={\it Mutate}($i_1$) & $i_2$={\it Initialize} & $i_2$={\it Initialize}\\
$i_3$={\it Mutate}($i_1$) & $i_3$={\it Select}($i_1, i_2$) & $i_3$={\it Mutate}($i_1$) & $i_3$={\it Mutate}($i_1$)\\
$i_4$={\it Select}($i_1, i_3$) & $i_4$={\it Mutate}($i_3$) & $i_4$={\it Select}($i_1,i_3$) & $i_4$={\it Select}($i_1,i_3$)\\
$i_5$={\it Crossover}($i_1,i_4$) & & $i_5$={\it Crossover}($i_2,i_4$) & $i_5$={\it Mutate}($i_4$)\\
 & & $i_6$={\it Mutate}($i_5$) & $i_6$={\it Crossover}($i_2,i_5$)\\
\hline
\end{tabular}
\end{center}
\end{table}
\setlength{\tabcolsep}{1.4pt}
\paragraph{Remarks:}

\begin{itemize}
\item[$(i)$]In our model the {\it Crossover} operator always generates a single offspring from two parents. The crossover operators generating two offspring may also be designed to fit our evolutionary model.
\item[$(ii)$]The {\it Select} operator acts as a binary tournament selection. The best out of two individuals is always accepted as the selection result.
\item[$(iii)$]	The {\it Initialize, Crossover and Mutate} operators are problem dependent. 
\end{itemize}

\section {Fitness assignment}
\label{fitness_s}

We have to compute the quality of each EA encoded in the chromosome in order to establish the fitness of a MEP individual. For this purpose each EA encoded in a MEP chromosome is run on the particular problem being solved.

Roughly speaking the fitness of a MEP individual is equal to the fitness of the best solution generated by one of the evolutionary algorithms encoded in that MEP chromosome. But, since the EAs encoded in a MEP chromosome use pseudo-random numbers it is likely that successive runs of the same EA generate completely different solutions. This stability problem is handled in the following manner: each EA encoded in a MEP chromosome is executed (run) more times and the fitness of a MEP chromosome is the average of the fitness of the best EA encoded in that chromosome over all runs. In all of the experiments performed in this paper each EA encoded into a MEP chromosome was run 200 times.

\section{Numerical experiments}
\label{experiments_s}

In this section, we evolve an EA for function optimization. For training purposes we use the Griewangk’s function \cite{Yao}.

Griewangk’s test function is defined by the equation (1).

\begin{equation}
f(x)=\frac{1}{4000}\sum_{i=1}^{n}x_{i}^{2}-\prod_{i=1}^{n}cos\left(\frac{x_i}{\sqrt{i}}\right) + 1
\end{equation}

The domain of definition is $[-500, 500]^n$. We use $n = 5$ in this paper. The optimal solution is $x_0$ = (0,\dots,0) and $f(x_0)=0$. Griewangk's test function has many widespread local minima which are regularly distributed.

An important issue concerns the representation of the solutions evolved by the EAs encoded in an MEP chromosome and the specific genetic operators used for this purpose. The solutions evolved by the EAs encoded in MEP chromosomes are represented by using real values \cite{Goldberg} (i.e. a chromosome of the second level EA is an array of real values). By initialization, a random point within the definition domain is generated. The convex crossover with $\alpha = \frac12$ and the Gaussian mutation with $\sigma = 0.5$ are used.

\subsection{Experiment 1}

\begin{figure}[ht]
\centerline{\includegraphics{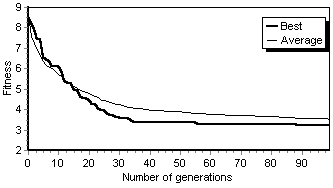}}
\caption{The fitness of the best individual in the best run and the average (over 10 runs) of the fitness of the best individual over all runs.}
\label{fig:Figure1}
\end{figure}

In this experiment we are interested in seeing the way in which the quality of the best evolved EA improves as the search process advances. The MEP algorithm parameters are: {\it Population size} = 100; {\it Code length} = 3000 genes; {\it Number of generations} = 100; {\it Crossover kind} = {\it Uniform}; {\it Crossover probability} = 0.7; {\it Mutations / chromosome} = 5; {\it Terminal set} = \{{\it Initialize}\}; {\it Function set} = \{{\it Select}, {\it Crossover}, {\it Mutate}\}. The results of this experiment are depicted in Fig.\ref{fig:Figure1}.

Fig.\ref{fig:Figure1} clearly shows the effectiveness of our approach. The MEP technique is able to evolve an EA for solving optimization problems. The quality of the best evolved EA is 8.5 at generation 0. That means that the fitness of the best solution obtained by the best evolved EA is 8.5 (averaged over 200 runs). This is a good result, knowing that the worst solution over the definition domain is about 313. After 100 generations the quality of the best evolved EA is 3.36.

\subsection{Experiment 2}

We are also interested in seeing how the structure of the best evolved EA changed during the search process.

\begin{figure}[ht]
\centerline{\includegraphics{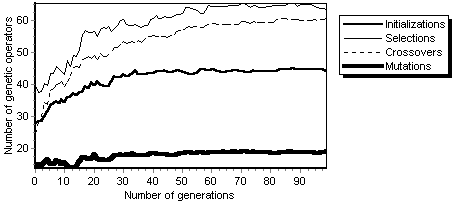}}
\caption{The fitness of the best individual in the best run and the average (over 10 runs) of the fitness of the best individual over all runs.}
\label{fig:Figure2}
\end{figure}

The evolution of the number of the genetic operators used by the best evolved EA is depicted in Fig.~\ref{fig:Figure2}.

From Fig.~\ref{fig:Figure2} it can be seen that the number of the genetic operators used by the best EA increases as the search process advances. For instance the averaged number of {\it Initializations} in the best EA from generation 0 is 27, while the averaged number of {\it Initializations} in the best evolved EA (after 100 generations) is 43. The averaged number of {\it Mutations} is small (less than 18) when compared to the number of occurrences of other genetic operators.

\section{Conclusions and further work}

A minutely described method for evolving evolutionary algorithms has been proposed in this paper. The numerical experiments emphasize the robustness and the effectiveness of this approach.

Further numerical experiments will analyze the relationship between the MEP parameters (such as the population size, the chromosome length, and the mutation probability) and the ability of the evolved EA to find optimal solutions. It is expected that an increased population size will bring about a substantial increase in the evolved EA performances.

The generalization ability of the evolved EA (how well it will perform on some new test problems) will also be studied. A larger set of functions should be used in order to increase the generalization ability.

An important issue is related to the amount of memory required by the evolved EA. No optimization regarding the memory used by the evolved EA was done in this paper. For instance, if the evolved EA performs 20 {\it Initializations}, 25 {it\ Selections}, 50 {\it Crossover} and 15 {it\ Mutations}, the memory required by the algorithm is 110 times the memory required for storing an individual. It is obvious that this amount of memory can be reduced because some memory locations can be overridden by newly created individuals. A simple algorithm that checks whether a memory location will be accessed or not in the future can be used for this purpose.

\end{document}